# Why Do We Need Neuro-Symbolic AI to Model Pragmatic Analogies?

Thilini Wijesiriwardene[1], Amit Sheth[1], Valerie L. Shalin[2], and Amitava Das[1]

[1]AI Institute, University of South Carolina, Columbia, SC, 29208, USA
[2]Wright State University, 3640 Colonel Glenn Hwy, Dayton, OH 45435, USA

A hallmark of intelligence is the ability to use a familiar domain to make inferences about a less familiar domain, known as analogical reasoning. In this article, we delve into the performance of Large Language Models (LLMs) in dealing with progressively complex analogies expressed in unstructured text. We discuss analogies at four distinct levels of complexity: lexical analogies, syntactic analogies, semantic analogies, and pragmatic analogies. As the analogies become complex, they require increasingly extensive, diverse knowledge beyond the textual content, unlikely to be found in the lexical co-occurrence statistics that power LLMs. To address this, we discuss the necessity of employing neurosymbolic AI techniques that combine statistical and symbolic AI, informing the representation of unstructured text to highlight and augment relevant content, provide abstraction and guide the mapping process. Our knowledge-informed approach maintains the efficiency of LLMs while preserving the ability to explain analogies for pedagogical applications.

## Introduction

The ability to reason analogically, using a familiar domain to make inferences about a new one, is fundamental to human cognitive ability [1, 2]. Analogies involve a familiar source domain and a less familiar target domain. Figure 1 presents a sample analogy between the familiar domain of cooking (source) to provide insight into the less familiar domain of photosynthesis by drawing comparisons and finding similarities between both. This analogical approach guides exploration and acquiring new understanding in the target domain.

The dominant contemporary approach to simulating intelligent behavior with Natural Language inputs is Large Language Modeling (LLM). LLMs combine billions of parameters with self-supervised learning over copious amounts of data to capture statistical regularities in the training data. LLMs excel at narrow, well-defined tasks in GLUE and SuperGLUE benchmarks. Given the substantial success of such neurally-inspired approaches in satisfactorily addressing narrow and well-defined tasks (e.g., classification, recommendation, and prediction), a growing number of leading AI researchers have identified new task goals, including abstraction and analogies (https://youtu.be/aeMbLkONLUw). We present a comprehensive analogy taxonomy and determine that LLMs can perform well on analogies at the taxonomy's lower, less complex levels. However, as we ascend the taxonomy, the analogies are more complex – what we call *Pragmatic Analogies* –specific, relatively rare knowledge outside of the text provides increasingly essential context. We advocate a Neuro-Symbolic approach (https://bit.ly/3KjbE68) informed not only by data but also relevant knowledge (https://bit.ly/DKduality), typically represented using Knowledge Graphs (KGs) (https://bit.ly/KGOKN). Below, we first present a taxonomy of analogies, highlighting the demands on knowledge outside of the analogy text to support analysis and the corresponding limitations of LLMs. Then, we sketch our neuro-symbolic approach, consistent with cognitive science but less concerned with fidelity to human processing, to focus on enriching fundamental computational methods with the required functionality.

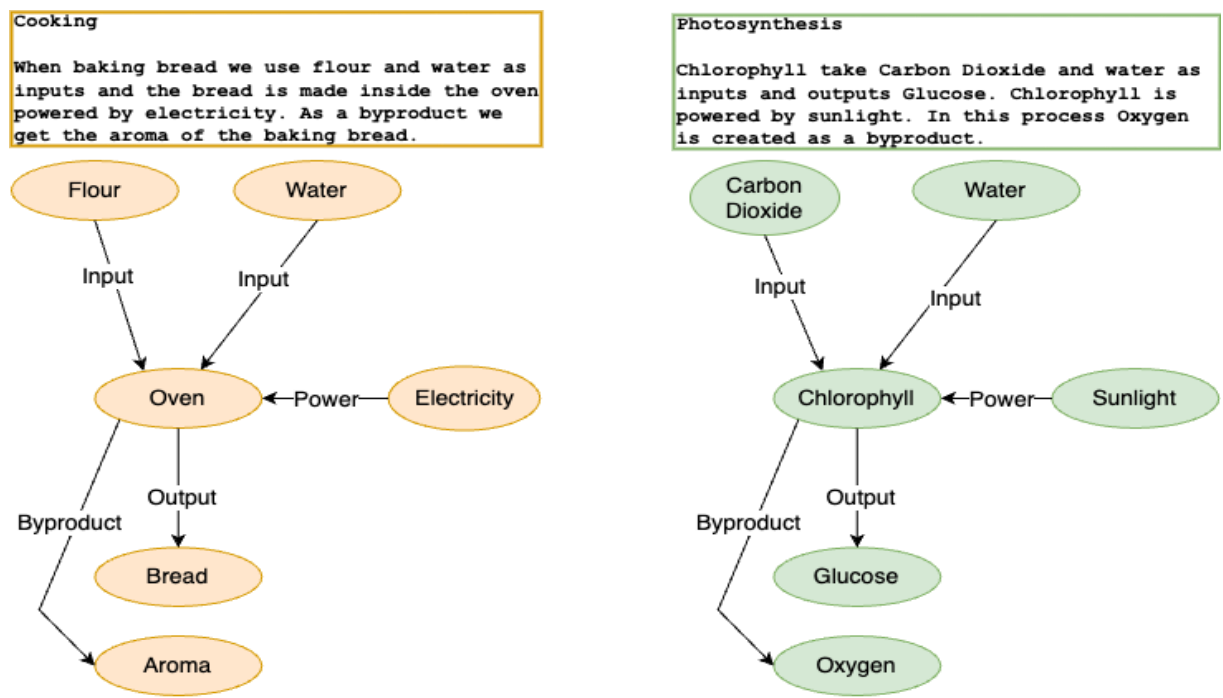

Figure 1: Analogy between source domain of cooking and target domain of photosynthesis.

# Analogies in Cognitive Science

A legacy cognitive science literature investigates how humans construct analogies. Built on the early psychometric studies of the four-word proportional analogy and supported by the prevalence of metaphor as the foundation of thought [a], Gentner, Hofstader, Holyoak, and their colleagues have led the field in analyzing rich multi-sentence descriptive analogies, generally separating the process into five steps: access, representation, mapping, inference, and schema establishment. All of these leaders aim to develop computational models of human analogical reasoning. Gentner and Forbus [b, c] focus primarily on the structural properties of analogy and the mapping process. Hofstader and Mitchell focus primarily on the initial representation of a proposed analogy, framing it as a constrained perceptual problem [d]. Similar to Hofstader, Holyoak and Thagard [e] emphasize the role of semantic and pragmatic constraints on problem representation, anticipating subsequent recognition of the centrality of causal reasoning [f]. Hofstader and Holyoak initially explored neurosymbolic computational models that combine neural mechanisms with symbolic knowledge, with Forbus examining the promise of neural mechanisms more recently [g]. Although less concerned with modeling human reasoning, common themes from the cognitive science work influence our own views. All cognitive science work acknowledges the need for external knowledge beyond the content in the analogy itself. All of them are concerned with the combinatorics of the mapping problem and the problem of evaluating necessarily imperfect analogy quality. Our own generally compatible effort focuses on representing unstructured, rich analogies, where external knowledge guides initial and layered representations associated with deep learning mechanisms.

## Types of Analogies

The past research on LLMs' abilities to model analogies primarily focused on simplistic proportional analogies found in the Miller Analogies Test for graduate school admissions and intelligence tests (e.g., "fingers is to hand as toes is to what?"). However, the LLM research community has recently considered richer analogies with multiple entities and complex relational structures typical of pedagogical and even creative settings. Identifying the limitations of LLMs for analogical reasoning requires a principled taxonomy. Curtis and Reigeluth [3] identified three types of textual analogies, ranging from simple analogies, where the base for the comparison between the source and target domain is not stated, to enriched analogies that include explanation. The third type is extended analogies, which include more potential relationships, some shared and others not. Here, we present a taxonomy based on [4] focused on the complexity and required external information and knowledge (Figure 2). Pragmatic Analogies are the most complex, spanning several sentences (often a paragraph) that elaborate on both the source and target domains, contain multiple concepts or entities related by diverse relationships, contain abstractions (modeled as subgraphs), and require us to map concepts/entities, relationships and subgraphs between source and target contextualized by external knowledge and a purpose, often pedagogical. Our taxonomy focuses on the complexity of the associated computation and the types and sources of external knowledge necessary for modeling each type.

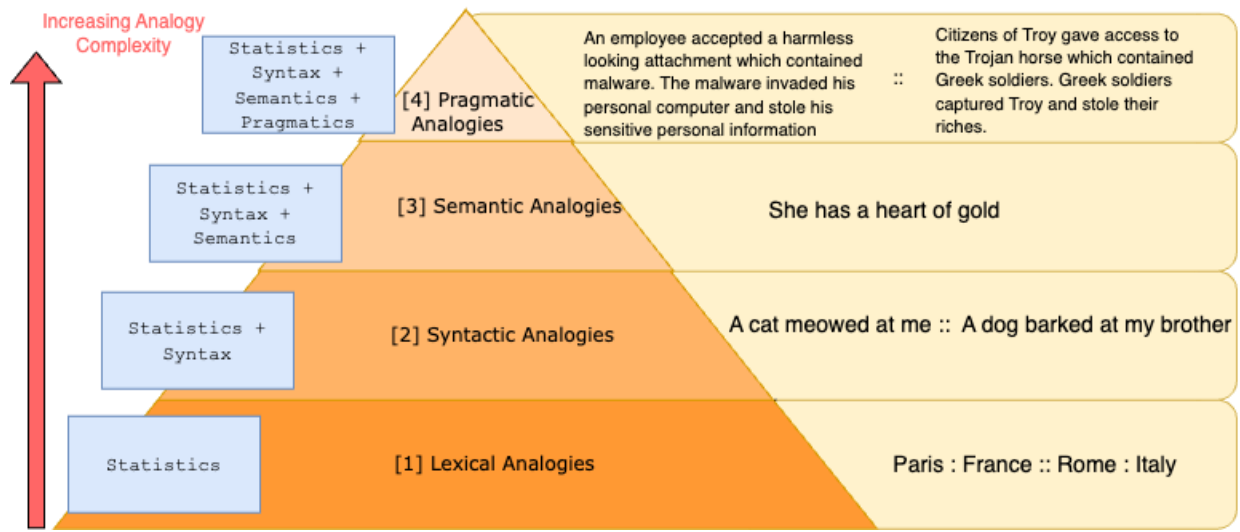

Figure 2: Taxonomy of analogies in order of increasing complexity. Examples for each level of the taxonomy are on the right, and the information/ knowledge needed to model analogies at that particular level are shown on the left.

## Lexical Analogies (Proportional Analogies)

The first level in Figure 2 consists of Lexical or Proportional Analogies. These primarily focus on lexical patterns across four terms linked by relational similarities that vary from explicit to implicit.

Proportional Analogies follow the format of a : b :: c : d, where the relationship between a and b, and the relationship between c and d, remains consistent and can either be explicit or implicit. For instance, when presented with "Paris : France :: Rome : ?," humans can easily retrieve the explicit relationship of "capital_of" from the first pair and apply it to the second pair to arrive at "Italy." However, the relationship between "Caricature : Drawing :: Limerick : ?." is unlikely to be pre-stored, but rather requires the more challenging construction of a relationship.

Explicit relationships such as 'type_of', 'capital_of', and 'part_of' are still possible for LLMs to tackle because the data fed into these models at training time likely includes these explicit and specific relationships. However, relationships beyond pre-existing categories are more challenging for LLMs. For example, the relationship between Caricature and Drawing does not fall into any pre-constructed hyponymy or hypernym categories. It might be loosely termed 'witty_creation' whereas 'type_of', 'capital_of', and 'part_of' relationships likely appear in the data as hyponym or hypernym relations.

> LLMs model Lexical Analogies effectively when they can depend on statistical regularities in data that inform the model. However, LLMs still struggle to capture rare and unusual relationships.

## Syntactic Analogies

Syntactic Analogies focus on the structural and grammatical similarities between phrases and examine word arrangement and grammatical relationships. The two lexical items involved in the analogy are compared to identify the similarities in their structural or grammatical patterns. These analogies are more complex than Lexical Analogies because they combine several words to form sentences, introducing relatively more complex relationships among the words, such as dependencies. For example, the sentences "A cat meowed at me" and "A dog barked at my brother" can be identified as analogous with similar syntactic and grammatical structures.

Rule-based techniques that parse grammatical and structural similarities can model Syntactic Analogies but with limited scalability. Current LLMs are well suited for the task, with their capability to reasonably capture simple syntactic and grammatical structures likely present in the training data [5], providing improved scalability compared to rule-based models.

## Semantic Analogies (Metaphors)

We identify an intermediate level of semantic analogies, largely expressed as a single proposition. Metaphor exemplifies this level. Consider "she has a heart of gold," comparing two unrelated domains: a person's anatomical attribute (heart) and the precious metal known for its value and purity (gold). Metaphor elements are distant from each other from a literal perspective, and the domains are blended rather than mapped [6].

Modeling such semantic analogies highlights the challenge of evaluating the understanding without the task-specific framings in levels 1 and 2, classically approached according to coherence, correspondence and connection to external knowledge. The statistical regularities, syntax, simple lexico-semantics [7], and grammatical structures captured by LLMs lack comprehensive knowledge about the semantics of the target and source domains. The inability of LLMs to distinguish between meaningful and nonsense assertions casts doubt on the LLMs' ability to address metaphors [8]. LLMs likely require additional mechanisms to access the necessary semantics.

## Pragmatic Analogies (Rich Analogies)

Pragmatic or Rich analogies employ longer narratives spanning multiple sentences, often a paragraph. Processing and comprehending such analogies require tackling compositionality, symbol manipulation and often commonsense causal knowledge.

Initial interpretation requires access to context beyond what is explicit in the text itself, particularly general knowledge of how these domains are structured and function independently, and some process for recovering the abstraction supporting the identification of relevant underlying patterns, principles, or attributes that can be translated and applied from one domain to another. Generic datasets are unlikely to contain lexical co-occurrences across domains, which is the foundation of LLMs [9]. The general consensus is that LLMs encounter difficulties when dealing with complex analogies that require more abstract mappings [4, 11]. Word order, critical to distinguishing roles, is only partially addressed with standard positional encoding methods [10]. Most critically, the structure necessary for mapping is distributed over lengthy textual descriptions and requires the apprehension of large-window dependencies using huge models and extended training. This approach is impractical for repeated encounters with pragmatic analogies. Complementary multi-faceted knowledge mitigates these challenges. In most domains, this knowledge is already curated by humans and made available via more generic KGs such as DBpedia, ConceptNet,

WordNet, Freebase and more domain-specific KGs such as Bio2RDF and Greek Mythology KG.

## **A Neuro-Symbolic Approach to Modeling Pragmatic Analogies**

Our technical approach fuses two complementary AI approaches: data-driven neural networks and knowledge-supported symbolic processing (https://bit.ly/3KjbE68). This fusion combines the efficiencies and scope of deep learning using large datasets with the ability to reason using explicit knowledge, usually represented as KGs.

We sketch our neuro-symbolic approach in two primary subsections: Analogy representation and mapping. In each subsection, we identify specific functions that require knowledge, including extracting and enriching concepts in the representation and distinct mapping tasks that are resolved with different computational methods.

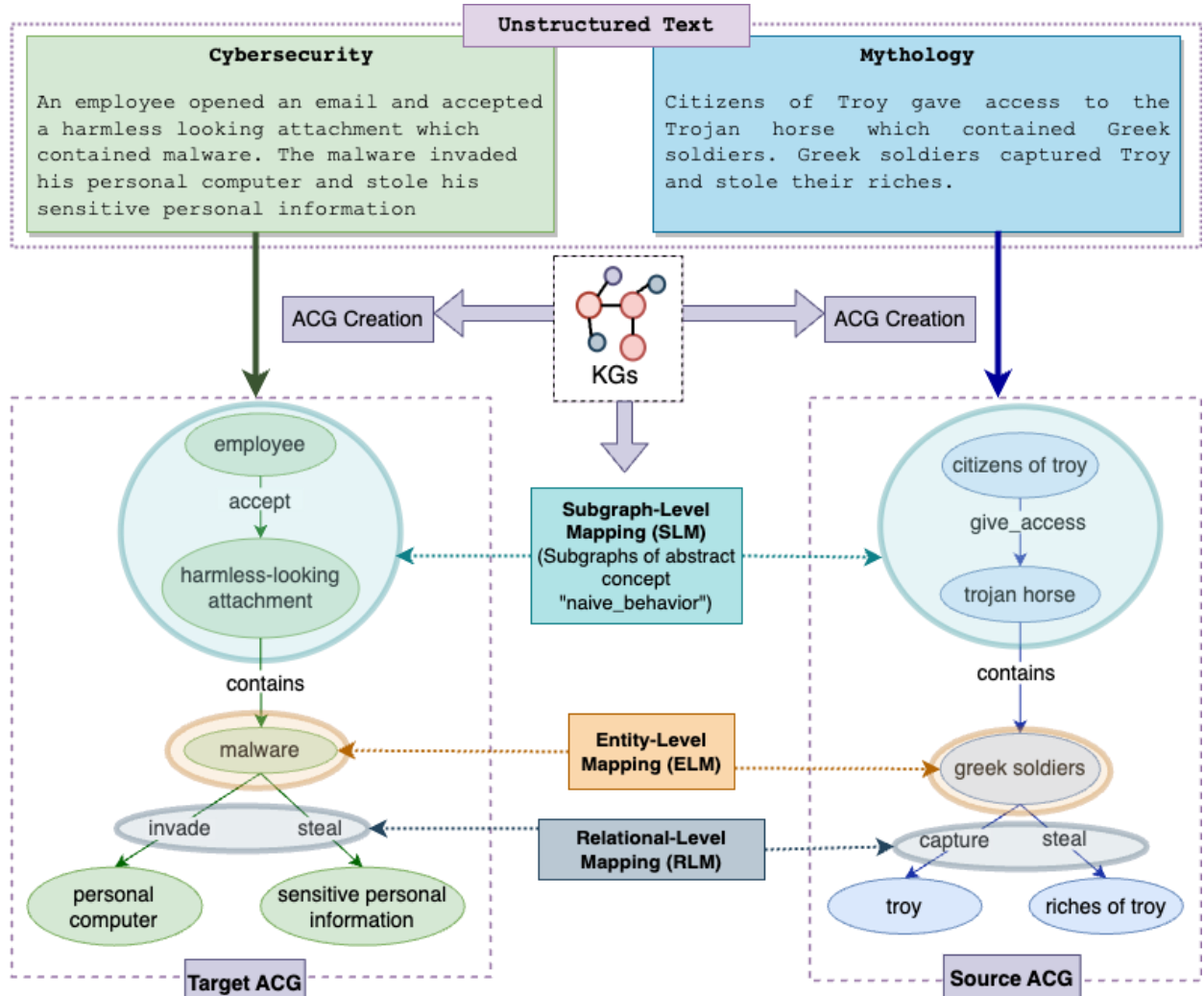

Figure 3: **Process of modeling Pragmatic Analogies:** Unstructured text of the target domain (Cybersecurity) and the source domain (Mythology) is parsed to create source

and target representations called Analogy Concept Graphs (ACGs). KGs are utilized in this process. ELM, RLM and SLM are performed after creating ACGs.

## Pragmatic Analogy Representation

We begin with unstructured text, consistent with the input to LLMs. However, we propose the recovery of a mediating graphical structure for the purpose of analogical reasoning, which we refer to as "Analogy Concept Graphs" (ACGs). In contrast to the immediate construction of embedding representations used in LLMs, ACG provides two critical functions. First, it captures the purpose of analysis, thereby directing subsequent processing. Second, it supports the recovery of an explanation for pedagogical applications.

Knowledge-based aid in identifying essential analogy concepts from the textual descriptions of both the source and target domains, which will be incorporated into the recovery of an ACG through a process we identify as "Analogy Concept Extraction." A subsequent process, ACG Enrichment, addresses the possibility that certain domain concepts may be missing in the textual descriptions.

### Analogy Concept Extraction (ACE)

Generating meaningful ACGs for both the source and target domains hinges on integrating world knowledge (also called the world model) to guide the extraction of essential and important concepts from the provided textual descriptions. Particularly for ACE, both domain-specific KGs and commonsense KGs are helpful.

Figure 3 showcases a target text containing various nouns and verbs. Some researchers may resort to rule-based methods or basic Named Entity Recognition models to extract essential concepts (entities and relationships) as triples from the textual descriptions. However, this simplistic extraction may yield triples that are not particularly pertinent for meaningful ACGs, such as <employee, opens, email>. The computation must focus on identifying triples that capture knowledge relevant to Cybersecurity, such as <harmless-looking attachment, contain, malware>, to address this challenge. We employ the Greek Mythology KG to score the extracted triples, a good example of how the purpose of analogy analysis may direct processing between disparate segments of text. These specialized KGs determine the relevance of each triple, selecting only those with high scores that hold crucial information for constructing the ACG. We note that specialized KGs are not always sufficient; we can also employ Commonsense KGs to score the triples extracted from more generic text in the source textual description.

**ACG Enrichment (ACGE)**

Knowledge participates in the ACG construction in another way, by filling in any gaps left by the initial text-based extraction. For example, the meaning of the extracted concept "invade" can be further enriched through WordNet, attaching a more comprehensive explanation, such as "penetrate or assault, in a harmful or injurious way." In this way, adding knowledge reduces the ambiguities between potential mappings, paradoxically reducing computational complexity. Enrichment can be done more expansively by incorporating complementary explanations from more than one relevant KG.

## **Pragmatic Analogy Mapping**

Mapping is a known challenge to computational approaches to analogical reasoning, due to computational complexity. We partition mapping into three different problems with three different computational solutions: Entity-Level Mapping, Relational-Level Mapping and Subgraph Level mapping.

**Entity-Level Mapping (ELM)**
Entities can be either named entities (in which case mapping is more specific and less abstract) or generic entities (in which case the mapping is slightly less specific and more abstract). We argue that ELM is relatively unimportant. Consider the following example. "citizens_of_troy" can be considered a named entity and "employee" as a generic entity. However, the mapping between these two entities does not give us any insight into the target domain, which is the Cybersecurity malware attack. Similarly, comparing "trojan_horse" and "harmless_looking_attachment" at the surface level does not provide any insights into how a malware attack could be similar to the trojan horse scenario in mythology. To make the connection, models need mappings beyond mere entity level.

**Relational-Level Mapping (RLM)**
Relational mapping is more abstract than entity mapping; two seemingly similar relations could mean different things in the source and target domains. Analogical relations are ordered. Also, the order of the relations needs to be identified, and mapping needs to reflect the relational order to be effective. Models need to refer to rich, extensive external knowledge represented in KGs to situate these abstract relations and relational orders, as LLMs cannot reliably capture such intricate semantics and pragmatics of language [12].

Relations are similar between the source and target domain (e.g., contain) in some cases and different in others (e.g., accepts vs. give_access). But in both situations, the relations can mean different things based on their domain, making the relations more abstract than entities. Also, the relational structure between the target and source domain must be ordered and aligned sequentially. To do this, the model must identify what each relation means in each domain and how they are ordered. This is particularly challenging for standalone LLMs because the semantic and pragmatic distances between the relations and relational structures are higher (mapping the same relational word in the source and target domain lexically focusing on the surface level is easier, but when the relational words are far apart semantically and pragmatically, this strategy can be less effective and erroneous).

**Subgraph-Level Mapping (SLM)**

SLM is the highest level of abstraction in analogical mapping. In SLM, the mapping is done on the subgraph level, which is more abstract than the two levels identified before. Subgraphs include a subset of entities and relationships and represent several abstract concepts that hold importance in source and target domains. For example, in Figure 3, the abstract concept of "naive_behaviour" is represented by both <employee, accepts, harmless-looking_software> in the target domain of Cybersecurity and <citizens_of_troy, give_access, trojen_horse> in the source domain of Greek mythology and eventually mapped (hierarchical concept representation in KGs support identifying these subgraphs). To identify that the specific Greek mythology includes the abstract concept of "naive_behavior," the model must utilize already synthesized knowledge present in relevant domain-specific KGs (in this specific case, to situate the meaning of trojen_horse and citizens_of_troy the model needs to reach out to domain-specific KG on greek mythologies). Current LLMs do not capture these types of high-level abstractions well.

Subgraph mappings require approximate Graph Isomorphism (GI). The task is challenging due to the NP-complete nature of GI. Various techniques for computing GI have emerged over time, encompassing backtracking algorithms, color refinement algorithms, group theoretic methods, Weisfeiler-Leman algorithm, Babai's quasi polynomial-time algorithm and Graph Neural Network (GNN) based methods. GNNs, as structure-driven models, dynamically adjust their network structure to capture intricate dependencies within an input graph. This results from an iterative aggregation process in GNNs, ultimately mapping the subgraph structure into a vector space. Once mapped onto a vector space, the similarity calculations can be done quite efficiently.

## **Parting Thoughts**

This article highlights different types of analogies based on their complexity and requirement of information and knowledge, guided by a four-level taxonomy. We argued that LLMs generally model simple analogies like Lexical Analogies and Syntactic Analogies fairly well, but they fall short when modeling Semantic and Pragmatic Analogies. We further explained how a neurosymbolic AI approach can be used to model Pragmatic Analogies by incorporating content beyond the text, capturing broad, rich, multi-faceted knowledge in the form of KGs. We discussed how ACGs could be created and enriched through the incorporation of KGs and how the mapping between source and target ACGs for three distinct levels of mapping, from Entity-Level Mapping to Relational-Level Mapping to Subgraph-Level Mapping. Our neurosymbolic approach supports explanations central to pedagogical applications.

> Modeling Pragmatic Analogies presents a formidable challenge for LLMs, as it requires an understanding of context that transcends mere statistics, syntax, and semantics. LLMs must be combined with rich, nuanced and multi-faceted KGs to acquire pragmatics.

**Acknowledgement**
The work in this article was supported in part by the National Science Foundation under Grant 2133842: "EAGER: Advancing Neuro-symbolic AI with Deep Knowledge-infused Learning."

Thilini is a Ph.D. student at the AI Institute of South Carolina (AIISC). Her research interests are neurosymbolic AI and NLU. Contact her at thilini@sc.edu.

Amit Sheth is the founding director of the AIISC, an NCR Chair and Professor of Computer Science & Engineering at the University of South Carolina. http://aiisc.ai/amit

Valerie L. Shalin is a Professor of Psychology at Wright State University, specializing in applications of technology to workplace activities. Contact her at valerie.shalin@wright.edu.

Amitava Das is a Research Associate Professor at the AIISC, USC. His research interests encompass NLP, multimodality, and social computing. http://amitavadas.com/